\documentclass[review]{elsarticle}

\usepackage{hyperref}

\usepackage{amsfonts}
\usepackage{epstopdf}

\usepackage{graphicx}
\usepackage{amsmath,amssymb} 
\usepackage{color}

\usepackage{multirow}
\usepackage{multicol}
\usepackage{threeparttable}
\usepackage{subfigure}
\usepackage{booktabs}
\usepackage{arydshln}
\usepackage{colortbl}
\usepackage{mathrsfs}
\usepackage{bbding}

\begin{document}
\begin{frontmatter}
	
	\title{Segmentation Mask Guided End-to-End Person Search} 
	
	
	
	\author[firstaddress]{Dingyuan Zheng}
	\author[firstaddress]{Jimin Xiao\corref{mycorrespondingauthor}}
	\cortext[mycorrespondingauthor]{Corresponding author}
	\ead{jimin.xiao@xjtlu.edu.cn}
	\author{Kaizhu Huang \fnref{firstaddress}}
	\author{Yao Zhao \fnref{secondaddress}}
	
	\address[firstaddress]{Department of Electrical and Electronic Engineering, Xi'an Jiaotong-Liverpool University, Suzhou, China}
	\address[secondaddress]{ Institute of Information Science, Beijing Jiaotong University, Beijing, China}
	
	\fntext[myfootnote]{D. Zheng, J. Xiao, K. Huang are with the Department of Electrical and Electronic Engineering, Xi’an Jiaotong-Liverpool University, Suzhou, China (e-mail: dingyuan.zheng, jimin.xiao,
		kaizhu.huang@xjtlu.edu.cn).}
	\fntext[myfootnote]{Y. Zhao is with Institute of Information Science, Beijing Jiaotong University, Beijing, China (e-mail: yzhao@bjtu.edu.cn).}
	
	
	
	\begin{abstract}
		Person search aims to search for a target person among multiple images recorded by multiple surveillance cameras, which faces various challenges from both pedestrian detection and person re-identification. Besides the large intra-class variations owing to various illumination conditions, occlusions and varying poses, background clutters in the detected pedestrian bounding boxes further deteriorate the extracted features for each person, making them less discriminative. To tackle these problems, we develop a novel approach which guides the network with segmentation masks so that discriminative features can be learned invariant to the background clutters. We demonstrate that joint optimization of pedestrian detection, person re-identification and pedestrian segmentation enables to produce more discriminative features for pedestrian, and consequently leads to better person search performance. Extensive experiments on benchmark dataset CUHK-SYSU, show that our proposed model achieves the state-of-the-art performance with 86.3\% mAP and 86.5\% top-1 accuracy respectively.
		\begin{keyword}
			person searh \sep re-identification \sep pedestrian detection \sep segmentation masks \sep background clutters
		\end{keyword}
	\end{abstract}
\end{frontmatter}

\section{Introduction}
Person re-identification has been widely applied in video surveillance systems with increasing demands in urban safety. It has attracted great attention in the computer vision community during the last decade. Person re-identification is generally solved as a retrieval problem \cite{zheng2016person, hermans2017defense}. Given a probe image, person re-identification aims to find all the images in the gallery set with the same identity. However, person re-identification has not be fully addressed, since the images captured by cameras are usually with the characteristics of illumination variations, occlusions and low resolution owing to the  shooting environment. These challenges potentially increase the intra-class variations and raise the recognition difficulty. 

\begin{figure}[ht]
	\centering
	\includegraphics[width=1 \linewidth]{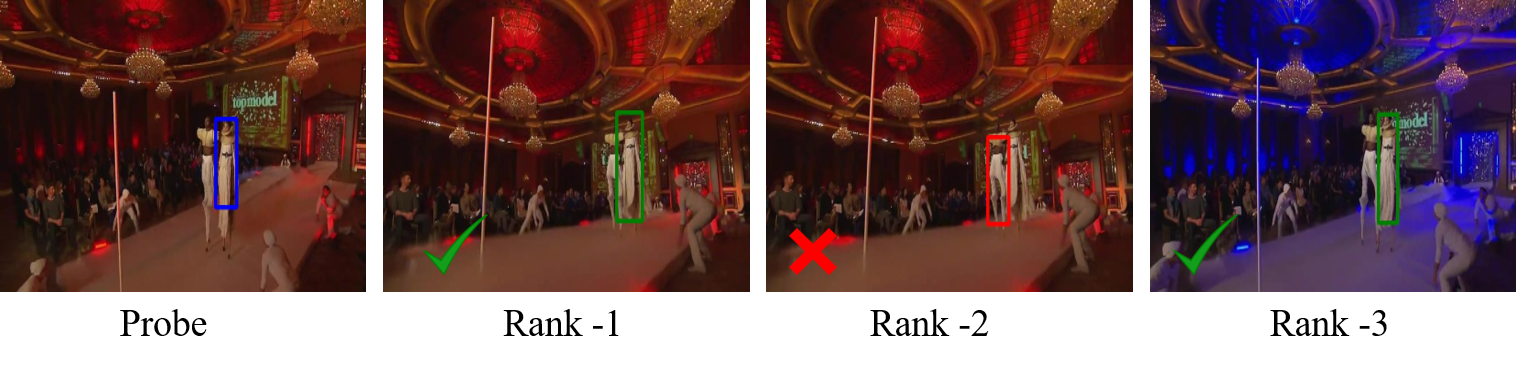}\\

	\setlength{\abovecaptionskip}{10pt}%
	\caption{One example to show the negative effect of background clutters which deteriorate the person re-identification performance. Blue box in the first column is the probe image. Other columns are the searching results from rank-1 to rank-3. Green boxes indicate the correct searching result, while the red box indicates the incorrect searching result.}
	
	\label{fig1}
\end{figure}

To this end, a great deal of research works on person re-identification devote to extract more discriminative features to represent human individuals, either by hand-crafted features \cite{liao2015person, gray2008viewpoint} or by CNN features \cite{wu2016personnet, wang2016joint}. Most of the existing person re-identification methods engage on cropped pedestrian bounding boxes without considering background clutters. Specifically, human individuals are represented by the features extracted from the regions constrained with the detected pedestrian bounding boxes, and Euclidean distance is computed to evaluate the similarity level among those probe-gallery pairs. It may result in a situation that different persons with similar background are close in the learned feature space. For example, in Fig. \ref{fig1}, the person in the bounding box in the third figure is different from the probe image. However, it is ranked before the person in the bonding box of the fourth figure who has the same identification as the probe image; this is simply because its background is more similar with the probe image.      

One straightforward yet effective solution to tackle the problem is to make the foreground part such as human body as the dominant region for feature extraction. In \cite{zhao2017spindle}, it adopts pose estimation approach to locate the key body points, and then aggregates the local features extracted from the pre-defined body regions with the global features extracted from the whole image. Based on the similar ideas, Tian \emph{et al.} propose a person-region guided pooling network with the assist of human parsing maps to solve the background bias problem \cite{tian2018eliminating}. Recently, 
researchers also attempt to introduce the attention mechanism into the person re-identification task for pedestrian feature extraction \cite{tay2019aanet, li2018diversity, li2018harmonious}.

Person search aims to search for the targeting person among multiple images recorded with different surveillance cameras, where the pedestrian bounding boxes are not available. Person search, different from person re-identification which assumes most of the pedestrian bounding boxes are manually cropped or perfectly detected by the state-of-the-art detectors, i.e. Faster R-CNN \cite{ren2015faster}, handles the challenges from both pedestrian detection and re-identification. Specifically, considering the step of pedestrian detection, the misalignment and false alarm caused by detectors further decrease the recognition rate \cite{ouyang2013joint, ouyang2012discriminative}. Meanwhile person search also has the aforementioned problem resulting from the background clutters in the generated pedestrian bounding boxes.      

In a recent work, Chen \emph{et al.} adopt segmentation mask to solve the background clutter problem in the person search task \cite{chen2018person}. Specifically, a two-stream model is established to extract the pedestrian features with one stream to emphasize the foreground information for the regions covered by the segmentation mask, and second stream to retain the global information for the original image. However, in \cite{chen2018person} the foreground regions are heuristically fixed annotation, in other words, to what extent the background should be removed is decided by the pedestrian segmentation mask. Besides, this work  separates the steps of pedestrian masking, pedestrian detection and person re-identification, which ignores the fact that jointly optimizing these steps can further bring in performance gain.            

Inspired by the previous works \cite{xiao2016end, xiao2017joint, xiao2019ian} that solve the person search task in an end-to-end manner, we propose an novel end-to-end person search framework that uses the segmentation mask to mitigate the negative effect of background clutters. Different from the previous work \cite{chen2018person} that designates the foreground regions by the segmentation masks explicitly, we utilize the segmentation mask to guide the feature extraction network to learn the enriched foreground features through a parallel mask branch. To do this, segmentation masks are precisely labeled in our new created dataset. Our proposed person search approach jointly optimizes pedestrian detection, person re-identification and pedestrian segmentation, which obtains more discriminative features for pedestrians benefiting from end-to-end learning. 

We summarize our contributions are as follows. 

\begin{itemize}
\item We propose a segmentation masks guided person search framework so as to mitigate  the negative effect of the background clutters in the detected pedestrian bounding boxes. Our proposed person search framework is trained end-to-end which considers the inherent relations among pedestrian detection, person re-identification and pedestrian segmentation, and hence more discrimitive features for pedestrians can be learned, which effectively enhance the person search performance.  
	
\item We create a new dataset which contains precise pedestrian segmentation mask annotations for 1,833 images from the existing CUHK-SYSU dataset. The dataset will be released for the future segmentation mask based person search research, which can be downloaded from the link: https://github.com/Dingyuan-Zheng/maskPS. Meanwhile, it is found that our approach only requires partial annotations for the segmentation masks rather than that for the whole dataset.
	
\item Extensive experiments on benchmark dataset CUHK-SYSU show that our proposed segmentation masks guided end-to-end person search framework outperforms a wide range of state-of-the-art person search methods, obtaining 86.3\% mAP and 86.5\% top-1 accuracy, respectively.
\end{itemize}




\section{Related Work}
In this section, we first review the existing works for the two sub-tasks in person search, pedestrian detection and person re-identification respectively. We then review the recent achievements on person search. 

\subsection{Pedestrian Detection}
Pedestrian detection has witnessed significant improvement in the past few decades. The first landmark work achieved by Dalal \emph{et al.}\cite{dalal2005histograms} adopts the architecture of HOG$+$SVM, and then DPM \cite{felzenszwalb2009object} is developed to better address the occlusion issue. After that, ICF \cite{dollar2009integral} and its variants \cite{zhang2014informed, nam2014local} outperform the previous hand-crafted feature based pedestrian detection methods. More recently, great progress has been made on the realm of general object detection benefiting from the convolutional neural networks \cite{filali2016multi, cholakkal2016classifier, dai2016r, lin2017feature, girshick2015fast, ren2015faster}. Further, \cite{zhang2016faster} discussed the feasibility of Faster R-CNN on pedestrian detection task. In this paper, we also adopt Faster R-CNN as our pedestrian detector.                

\subsection{Person Re-identification}
With the great success of convolutional neural networks, researchers have proposed numerous deep learning based person re-identification solutions \cite{zheng2017person, cheng2016person, varior2016gated, xiao2016learning, mclaughlin2016recurrent}. The re-identification system is typically composed of two categories, feature extraction and similarity metrics learning. Some researchers attempt to improve the person re-identification performance by taking the advantage of enhanced feature representation. For instance, in \cite{yi2014deep, li2014deepreid}, the original image is horizontally split into patches, and part matching is then applied among these generated local patches. In \cite{zhao2017spindle}, local features of the body sub-regions defined by the pose estimation results are merged with the whole body features to improve the robustness of the final feature representation. Other researchers propose better person re-identification solutions by using well-designed similarity metrics learning. Generally, one category adopts verification loss, for example, contrastive loss \cite{varior2016gated}, triplet loss \cite{liu2017end} or quadruplet loss \cite{chen2017beyond}, while another category utilizes identification loss \cite{zheng2017person, zheng2016mars} or both \cite{geng2016deep}. In this paper, our person search framework is built upon the identification model.    

\subsection{Person Search}
As an extension of the conventional person re-identification, person search retrieves the target person from the raw scene images, where pedestrian bounding boxes are not available \cite{xu2014person}. In the pioneer work \cite{xiao2017joint}, Xiao \emph{et al.} show that pedestrian detection and person identification could be solved in an end-to-end framework. Following this work, Xiao \emph{et al.} \cite{xiao2019ian} enhance the discriminability of the pedestrian features by introducing center loss. Liu \emph{et al.} \cite{liu2017neural} recursively shrink the attentive regions till the target person is retrieved. In \cite{munjal2019query}, global context of query-gallery pairs are emphasized by establishing a query-guided region proposal network and similarity sub-network in a siamese structure. Yan \emph{et al.} further improve the person search performance by exploiting co-travelers as global context. A recent person search approach \cite{chen2018person} uses segmentation mask to filter the foreground person from the original input and aggregates the features of both foreground and whole image which are extracted from a two-stream model. Besides, the authors also state that better person search performance can be achieved by solving pedestrian detection and identification separately with off-line pedestrian masks. Different from \cite{chen2018person}, we optimize jointly these three tasks in an end-to-end framework. In particular, we use the segmentation mask to guide the network to learn the discriminative regions automatically rather than explicitly specifying these regions.

\section{Proposed Method}

In this section, we propose a novel partially labeled segmentation masks guided person search framework, as shown in Fig. \ref{fig3}. We first introduce our new dataset which contains partially labeled segmentation masks, and then elaborate our end-to-end person search framework.   
\begin{figure}[ht!]
	\centering
	\subfigure[]{
		\includegraphics[width=0.22\textwidth]{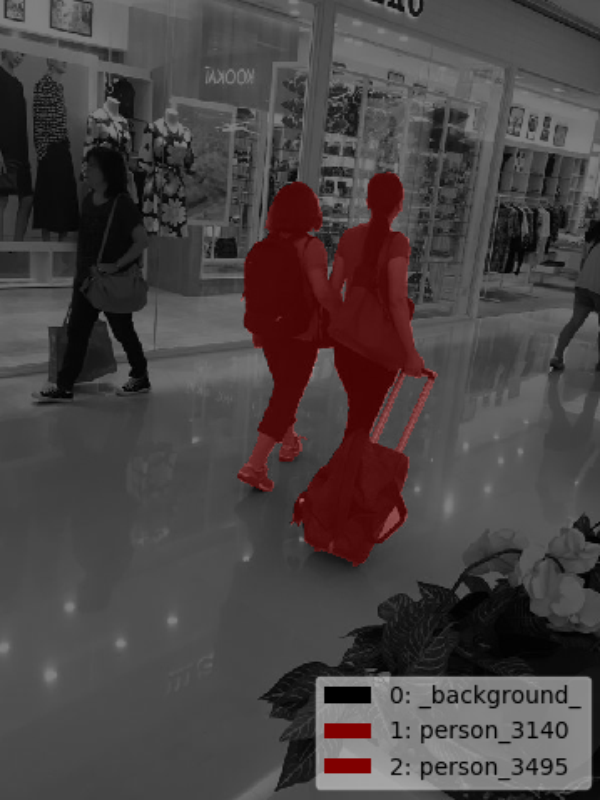}}
	\subfigure[]{
		\includegraphics[width=0.22\textwidth]{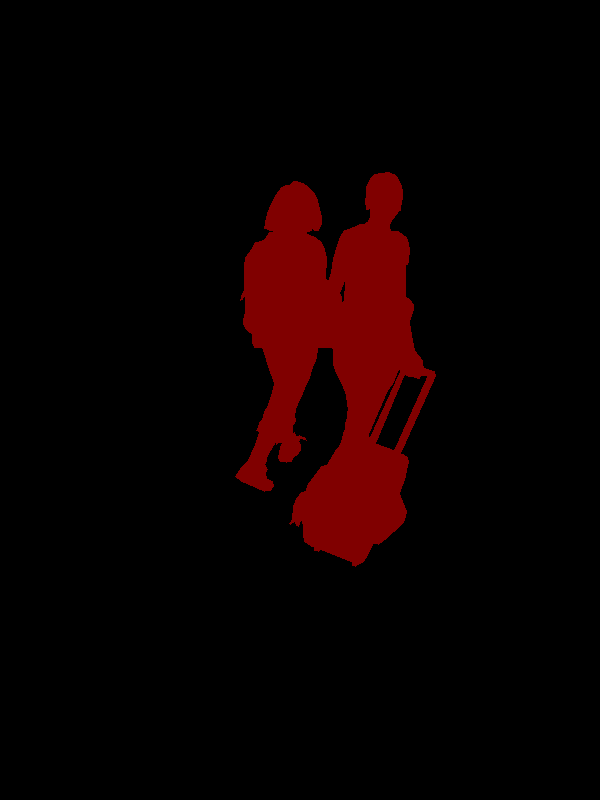}}
	\subfigure[]{
		\includegraphics[width=0.22\textwidth]{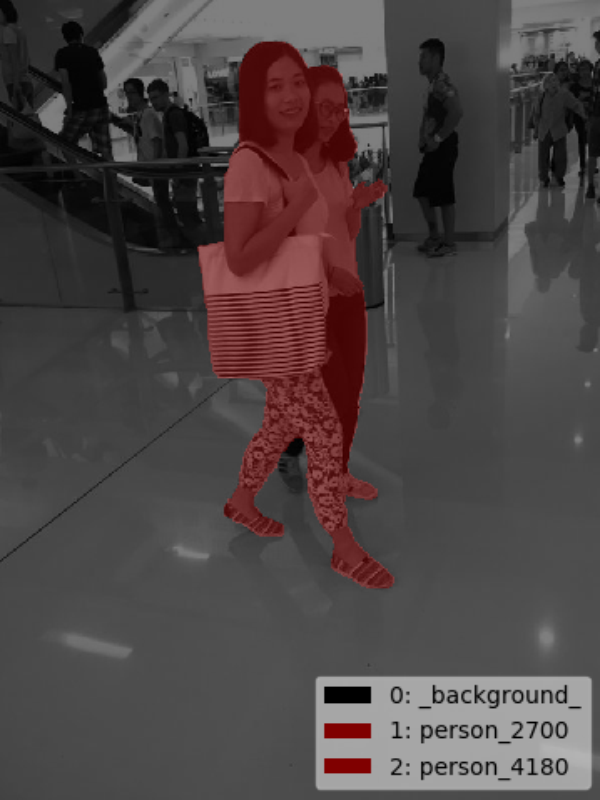}}
	\subfigure[]{
		\includegraphics[width=0.22\textwidth]{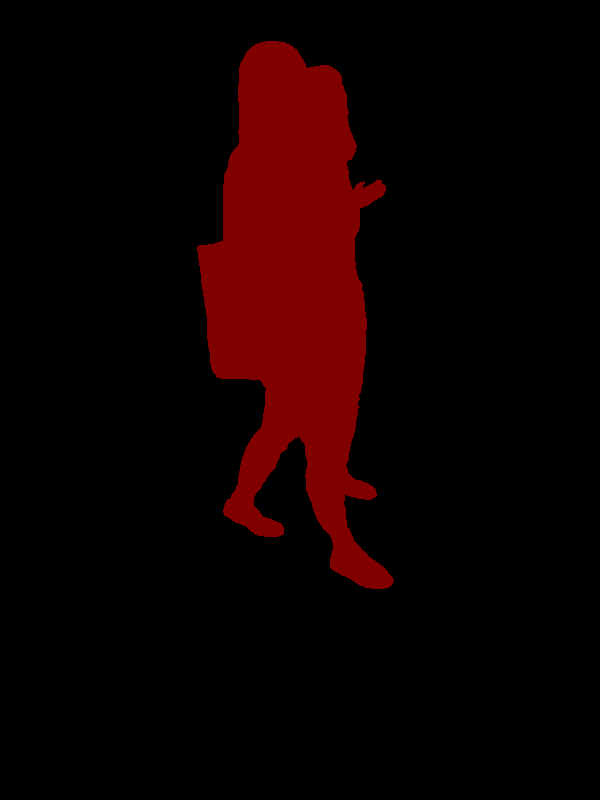}}
	
	\setlength{\abovecaptionskip}{10pt}%
	\caption{Two examples in our created dataset. We only provide the segmentation masks for the labeled persons (the persons labeled with [1-5532] in CUHK-SYSU dataset). The shadow regions in first and third columns indicate the labeled persons. The second and fourth columns are their segmentation masks.} 
	
	\label{fig2}
\end{figure}

\subsection{A New Dataset with Partially Labeled Segmentation Masks}

To the best of our knowledge, all current segmentation masks based person search/re-ID approaches are based on off-line pedestrian masks generated from Fully Convolutional Networks (FCN) \cite{long2015fully} or Fully Convolutional Instance-aware Semantic Segmentation (FCIS) \cite{li2017fully} without considering the benefit from joint optimization of pedestrian segmentation and person re-identification tasks. To extract more discriminative features and mitigate the negative effect of background clutters as well as to build an end-to-end framework that jointly optimize pedestrian detection, person identification and pedestrian segmentation, we created a new person search dataset to provide the precise annotations of pedestrian segmentation masks. We labeled the pedestrian segmentation masks for a portion of images in CUHK-SYSU dataset \cite{xiao2016end}. 

CUHK-SYSU dataset \cite{xiao2016end} is a large-scale dataset for person search, and the data is collected from diverse scenes. Specifically, it contains 18,184 images, 6,057 query persons in 12,490 images are captured from the street, while the rest 2,375 query persons in 5,694 images are collected from the movies and dramas. The dataset is split into the training set and test set, and guarantees no overlap occurs on images and query persons between the training split and test split. Training split contains 11,206 images with 5,532 query persons, while test split contains 6,978 images with 2,900 query persons. Each query person appears in at least two images. The dataset also contains two subsets to evaluate the person search performance under low resolution and occlusion. The person identities of training split are in the range of [-1, 5532], where -1 indicates the unlabeled persons and 0 indicates the background (non-person). 

In our dataset, to guarantee the uniformity of data distribution, we divide the training set into \emph{N} portions (\emph{N}=2,241 in our case), with 5 images in each portion (except the last portion, which contains 6 images), and we randomly select one image from each portion, and filter out the images with only unlabeled persons (the persons labeled with -1). Finally 1,833 images in the training set are selected for the segmentation mask labeling. 

To the best of our knowledge, accessories, i.e, handbags, luggage cases and baby carriage, might act as suggestive context in person re-identification. In a consequence, we treat these objects as foreground during the mask annotating process. It should be noticed that we provide the mask annotations for only the labeled persons (the persons labeled with [1$\sim$5532]) in a raw scene image. The samples of the image with segmentation masks are shown in Fig. \ref{fig2}.

We utilize Labelme \cite{labelme2016} as the annotation tool. All our segmentation masks follow the unified annotation rules. When a person is occluded by non-person objects, we only keep the visible part of the occluded person, and the accessories are kept as well. We also give the statistics for our created dataset, as shown in Table \ref{tab1}. The selected 1,833 images from the CHUK-SYSU training set contain 9,084 pedestrians in total, with 2,815 labeled persons and 6,269 unlabeled persons. In particular, we only annotated the segmentation masks for the labeled persons. The rest 9,373 images in the training set contains 12,270 labeled persons and 33,918 unlabeled persons. 


\begin{table}
	\centering
	\begin{tabular}{p{2.7cm} p{3cm}<{\centering} p{1cm}<{\centering} p{3.5cm}<{\centering}}
		\hline
		Dataset & Number of images &  & Number of pedestrians \\
		\hline
		\multirow{2}{*}{Images with masks}& \multirow{2}{*}{1,833} &LP& \textbf{2,815} \\
		\cmidrule{4-4}		
		& &UP&6,269 \\
		\cdashline{1-4}
		\multirow{2}{*}{Images without masks}& \multirow{2}{*}{9,373} &LP& 12,270 \\
		\cmidrule{4-4}
		& &UP&33,918 \\
		\hline
	\end{tabular}
	\setlength{\abovecaptionskip}{10pt}%
	\caption{Statistics of our created dataset. LP: Labeled persons, UP: Unlabeled persons. The labeled persons (2,815) in the selected 1,833 images are annotated with pedestrian segmentation masks.}
	\label{tab1}
\end{table}

\subsection{Our Proposed Person Search Framework}
\begin{figure}[t]
	\centering
	\includegraphics[width=1\linewidth]{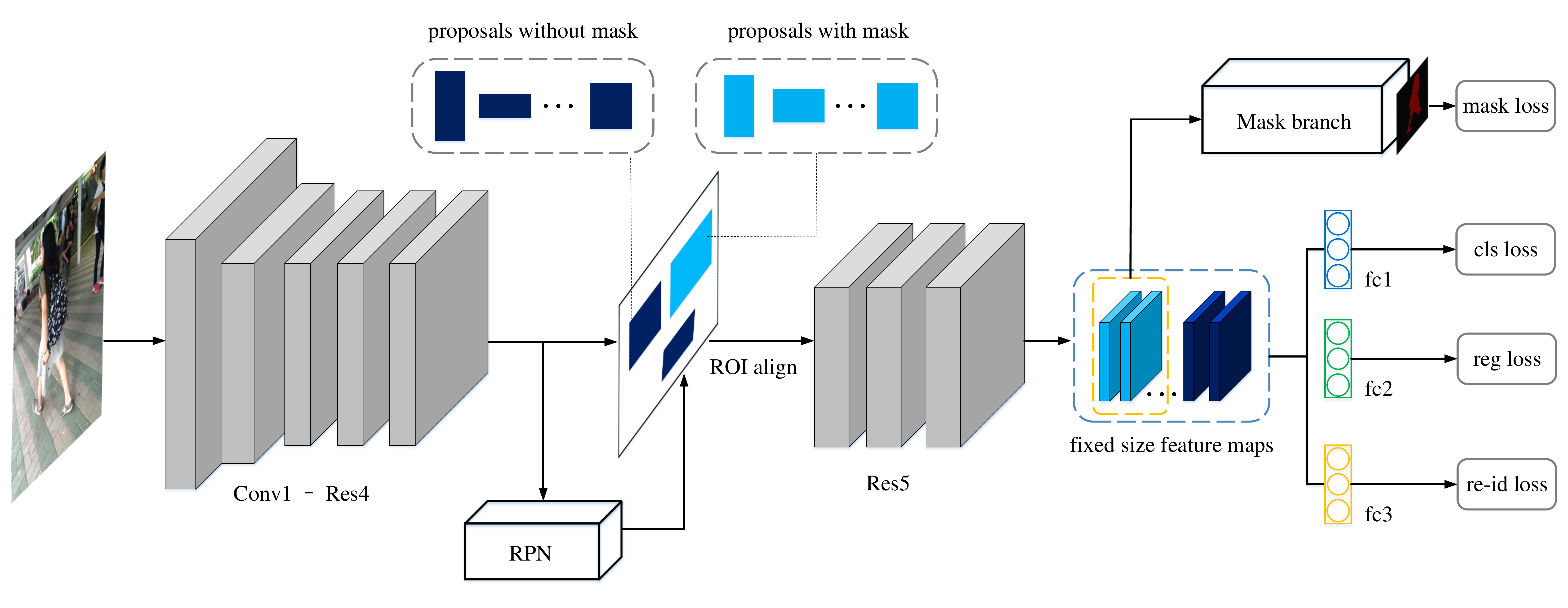}\\

	\setlength{\abovecaptionskip}{10pt}%
	\caption{The schematic of our proposed segmentation masks guided person search framework. The model is trained end-to-end with multi-task loss. We adopt RPN to generate candidate boxes, and we denote the proposals associated with labeled person as proposals with mask, while the proposals associated with unlabeled person as proposals without mask since we only partially label the segmentation masks for the labeled persons in a raw image. Feature vectors of all candidate boxes go into the regression loss, classification loss and identification loss, whereas only the feature maps of the proposals with mask are fed into the mask branch.} 
	
	\label{fig3}
\end{figure}

Person search aims to retrieve the target person across raw scene images without pedestrian bounding boxes. Our proposed approach jointly optimizes three sub-tasks including pedestrian detection, person identification and pedestrian segmentation in an end-to-end person search framework. Apart from the pedestrian detection module to produce online pedestrian bounding boxes and person identification module to categorize person identities, we further establish a parallel pedestrian segmentation branch to predict pedestrian masks. Benefiting from the end-to-end optimization of three tasks, more discriminative pedestrian features can be extracted. The overall schematic of the proposed segmentation masks guided end-to-end person search framework is shown in Fig. \ref{fig3}. The network is elaborated as follows. 

Arbitrary size images are resized such that the shorter side has 600 pixels. An image is then fed into the first part of the residual backbone network \cite{he2016deep}. Specifically, we divide the residual network into two parts, i.e, for ResNet-50, the first part contains the layers from Conv1 to Res4, and the rest Res5 forms the second part. 

To address pedestrian detection, we adopt the region proposal network \cite{ren2015faster} (RPN)  to produce online pedestrian proposals. RPN is trained with cross entropy loss to distinguish pedestrians and background, we express it as $\mathcal{L}_{cls}$:

\begin{equation}
\mathcal{L}_{cls} = -\sum_{i=1}^{N}y_ilog(s_i),
\end{equation}

\noindent where $N$ is the number of the generated proposals, $s_i$ is the prediction score and $y_i$ is the related ground truth which indicates person or non-person. We use the Smoothed-L1 loss \cite{girshick2015fast}, $\mathcal{L}_{reg}$, to regress the precise location for each pedestrian, as defined as follows: 

\begin{equation}
\label{regressionloss}
\mathcal{L}_{reg}=\left\{
\begin{array}{lcl}

0.5\mathcal{D}^2               &  & \vert{\mathcal{D}}\vert<1\\
\vert{\mathcal{D}}\vert-0.5    &  & otherwise,\\

\end{array} \right.
\end{equation}

\noindent where $\mathcal{D}$ denotes the coordinate differences between the predicted box and its related ground truth location, and these two losses together are denoted as $\mathcal{L}_{RPN}$. The generated candidate boxes are either associated with background or a foreground part (the ground truth bounding boxes). Since we only provide the mask annotation for the labeled persons (persons labeled with [1$\sim$5532]) in a raw scene image, those generated candidate boxes associated with foreground parts are consequently divided into two types. The first is the candidate boxes associated with labeled persons, and it is denoted as proposals with mask, while the second is the candidate boxes associated with unlabeled persons, which we denote as proposals without mask, as shown in Fig. \ref{fig3}.

All the proposals generated from RPN and the feature maps generated from the first part of the residual network are input into the ROIAlign layer \cite{he2017mask} to produce the fixed size feature map for each ROI.


Targeting for person identification, once the fixed size feature maps are obtained, these feature maps are further convolved into the second part of the residual network and the output, $\mathcal{F}_p\in\mathcal{R}^{c\times m\times m}$, are summarized into 2,048 dimensional feature vectors $f_p\in\mathcal{R}^c$ through an average pooling layer. Here $c$ is the channel width and $m$ denotes the size of the feature maps. To further reduce the false alarm caused by RPN and refine the predicted locations of the candidate pedestrians, $f_
p$ are then fed into two fully connected layers respectively and again supervised by $\mathcal{L}_{reg}$ and $\mathcal{L}_{cls}$ losses. Following \cite{xiao2017joint}, we denoted these two losses together as $\mathcal{L}_{RCNN}$. Besides, $f_p$ is projected into a 256 dimensional feature vector $f_{id}\in\mathcal{R}^d$ through the third fully connected layer followed by L2-normalization, which is used as the final feature representation for each retrieved pedestrian. In the training phase, we adopt OIM loss \cite{xiao2017joint} to supervise the person identification module, where $p_{id}$ indicates the probability of the identification features, $f_{id}$, belonging to $id$-th class,   


\begin{equation}
p_{id} = \frac{exp(v_{id}^Tf_{id})/\tau}{\sum_{j=1}^Lexp(v_j^Tf_{id}/\tau)+
	\sum_{k=1}^Qexp(u_k^Tf_{id}/\tau)}.\qquad
\end{equation}

\noindent Here $\tau$ is a parameter to control the softness of the probability function. The features of the labeled identities are stored in a lookup table with dimension \emph{L}, with $v_{id}$ denoting the current feature for class $id$ among 5,532 categories, and it is continuously updated during the training phase as follows:

\begin{equation}
v_{id} = \beta v_{id} + (1- \beta) f_{id},
\end{equation}   

\noindent where $\beta$ is a momentum parameter used to adjust the update rate. While the features of the unlabeled persons are stored in a circular queue with dimension $Q$, and $u_k$ indicates the features for the $k$-th unlabeled person. The objective of OIM loss is to maximize the expected log-likelihood, and the identification loss is then defined as:

\begin{equation}
\mathcal{L}_{identification} = E_x[logp_{id}].
\end{equation}

Most importantly, in order to improve the discriminability of $f_{id}$, we establish a parallel mask branch on top of the shared features $\mathcal{F}_p$. Specifically, we pick out the feature maps associated with labeled persons from $\mathcal{F}_p$, and use these feature maps $\mathcal{F}_{pm}\in\mathcal{R}^{c\times m\times m}$ to predict segmentation masks with the size of $2m \times 2m$ for each proposal with mask ($m$ equals 7 in our case), and the predicted masks are then computed into binary cross entropy loss \cite{he2017mask}, which can be written as:

\begin{equation}
\mathcal{L}_{mask} = 
\sum_{s=1}^{S}\sum_{t=1}^{T}(-Y_tlog(x_t)-(1-Y_t)log(1-x_t)),
\end{equation}

\noindent where $x_t$ denotes the probability of $t$-th pixel in the predicted mask being recognized as foreground, $Y_t$ is its related ground truth, $T$ is the number of pixels in the predicted pedestrian mask ($T=2m\times2m$), and $S$ is the number of proposals with mask.    

Finally, we adopt a multi-task loss to train our person search framework in an end-to-end manner. The total loss is defined as: 

\begin{equation}
\label{totalloss}
\mathcal{L}_{total} = \mathcal{L}_{RPN} + \mathcal{L}_{RCNN} + \mathcal{L}_{identification} + \lambda \mathcal{L}_{mask},
\end{equation}         


\noindent specifically, $\lambda = 1$ when input image contains labeled segmentation masks, otherwise, $\lambda = 0$.

With the assist of the partially labeled segmentation masks, our proposed person search framework can generate more discriminative features invariant to background clutters, compared with the previous segmentation mask based state-of-the-art approach \cite{chen2018person}.               

\section{Experimental Results} 
In this section, the dataset and evaluation metrics we used are first introduced, followed by implementation details. We also compare our proposed method with previous state-of-the-art results. At last, our proposed person search framework is verified in the ablation study. 



\subsection{Dataset and Evaluation Metrics} 
We use our newly labeled CHUK-SYSU dataset, as introduced in Sec. 3.1, in our experiments. We adopt both mean average precision (mAP) and top-1 matching rate to evaluate all our experiment performance, similar to \cite{xiao2017joint}. A matching is accepted only if the overlap between the detected pedestrian bounding boxes and the ground truth bounding boxes is larger than pre-defined intersection over union (IOU) threshold, which equals to $0.5$. 

\subsection{Implementation Details}
\textbf{Training Phase.} During training, we initialize our residual backbone with the ImageNet pretrained ResNet-50 model and adopt SGD as optimizer. The initial learning rate sets to $0.0004$ and decayed by a factor of $0.1$ at every $4$ epochs. Because of the large memory consumption of the Faster R-CNN framework \cite{ren2015faster}, we set the batch size to $1$ during the $12$ training epochs. All our experiments are implemented by Pytorch on Titan X Pascal GPU. 

It should be noticed that, the shorter side of the input images is resized to $600$ pixels, and we also augment the training data by horizontal flipping the training images and their related ground truth bounding boxes as well as the ground truth masks.
In paticular, the ground truth masks are resized to $14 \times 14$ to match the masks generated from the mask branch. For the implementation of the mask branch, we adopt a similar architecture as in the Mask R-CNN \cite{he2017mask}. $6$ losses are used jointly to supervise the training process. 

\textbf{Inference Phase.} At test time, the shorter side for both query and gallery images is resized to $600$ pixels as in the training process. We use the features generated from the last residual block (Res5) to represent each pedestrian, either probe person or the persons detected from the gallery set. Euclidean distance is then computed for each probe and gallery pair to assess the similarity level.

\subsection{Comparison with State-of-the-Art Approaches}

In this subsection, we report the person search performance of our model on our newly labeled person search dataset CUHK-SYSU, and we also give the comparison to several state-of-the-art approaches, including methods that optimize pedestrian detection and identification jointly (OIM \cite{xiao2017joint}, IAN \cite{xiao2019ian}, NPSM \cite{liu2017neural}, QEEPS \cite{munjal2019query} and GCNPS \cite{yan2019learning}), as well as the methods solving pedestrian detection and person identification separately (DSIFT+Euclidean\cite{zhao2013unsupervised}, DSIFT+KISS-\\ME\cite{koestinger2012large}, BoW\cite{zheng2015scalable}+Cosine similarity, LO-MO+XQDA, and MGTS \cite{chen2018person}). 

\subsubsection{Overall Person Search Performance on CUHK-SYSU}

The comparative results with gallery size $100$ are summarized in Table \ref{tab2}. We follow the annotations defined in \cite{chen2018person} and \cite{yan2019learning}, where ``CNN" denotes the Faster R-CNN detector with ResNet-50 backbone, and ``CNN$_v$" denotes the VGG-based detector. 

The methods above the dash line handle pedestrian detection and person identification separately. It can be observed that the deep CNN based pedestrian features \cite{chen2018person} achieved better performance than hand-crafted features \cite{zhao2013unsupervised}\cite{koestinger2012large}\cite{zheng2015scalable}. CNN$_v$+MGTS \cite{chen2018person} also utilizes segmentation mask to produce more discriminative features by filtering out the background, and achieves the best performance among those methods addressing pedestrian detection and person identification separately. Our proposed method uses segmentation mask to guide the network to extract discriminative pedestrian features by specifying the foreground regions. Meanwhile, pedestrian detection and person identification are optimized jointly. Our framework achieved 3\% gain compared with \cite{chen2018person} on both mAP and top-1 matching rate. 

All the joint methods (below the dash line) are built upon the Faster R-CNN \cite{ren2015faster} framework where OIM \cite{xiao2017joint} can be regarded as the benchmark.
The major distinction between our method and OIM \cite{xiao2017joint} is that a new pedestrian segmentation mask branch is added. We achieve a significant performance improvement, with 10.8\% mAP and 7.8\% top-1 higher compared with \cite{xiao2017joint}. It demonstrates the importance of the pedestrian segmentation mask and the newly labeled dataset. Other methods \cite{xiao2019ian}\cite{liu2017neural}\cite{munjal2019query}\cite{yan2019learning} are state-of-the-art person search approaches with good performance. IAN \cite{xiao2019ian} improves the person search performance by introducing center loss to reduce the intra-class variations. NPSM \cite{liu2017neural} designs a person search approach by recursively shrinking the search area. QEEPS \cite{munjal2019query} proposes a strong person search framework by learning query-guided global context. \cite{yan2019learning} utilizes GCN to explore the impact of context persons on the person search task. Nevertheless, we still achieve 2\% gain on both mAP and top-1 accuracy compared with \cite{munjal2019query}, and 2\% improvement on mAP compared with \cite{yan2019learning}, all of which prove the effectiveness of our method.

The visualization of person search results on the CUHK-SYSU dataset are shown in  Fig. \ref{fig5}. The upper images in each group are the searching results of OIM \cite{xiao2017joint}, and the lower images in each group are that of our model.  
It is observed that the persons in the bounding boxes of the third and the fourth images in the upper rows of group (a) and (b), as well as the person in the bounding box of the third image in the upper row of group (c), are different from their probe images. However, these persons are ranked before the persons who have the same identities as the probe images, simply because their background is more similar to the probe images. Nevertheless, with the assist of partially labeled segmentation masks, our model focus on the foreground and can distinguish persons based on the detailed textural information rather than the background-noise.

\begin{table}
	\centering
	\begin{tabular}{l c c}
		\hline
		\textbf{Method}            & \textbf{mAP(\%)}  & \textbf{top-1(\%)}\\
		\hline
		CNN + DSIFT + Euclidean \cite{zhao2013unsupervised} &34.5       & 39.4             \\
		CNN + DSIFT + KISSME \cite{zhao2013unsupervised}\cite{koestinger2012large}  & 47.8   &     53.6              \\
		CNN + BoW + Cosine \cite{zheng2015scalable}& 56.9              & 62.3              \\
		CNN + LOMO + XQDA \cite{liao2015person}  & 68.9              & 74.1              \\
		CNN$_v$ + MGTS \cite{chen2018person}               & 83.0              & 83.7              \\
		\cdashline{1-3}
		OIM \cite{xiao2017joint}                & 75.5              & 78.7              \\
		IAN(ResNet-50) \cite{xiao2019ian}     & 76.3              & 80.1              \\
		NPSM \cite{liu2017neural}               & 77.9              & 81.2              \\
		QEEPS \cite{munjal2019query}              & 84.4              & 84.4              \\
		GCNPS \cite{yan2019learning}               & 84.1              & \textbf{86.5}              \\
		\hline
		Ours                       & \textbf{86.3}           & \textbf{86.5}            \\  
		\hline
	\end{tabular}
	\setlength{\abovecaptionskip}{10pt}%
	\caption{Comparison with the state-of-the-art on CUHK-SYSU dataset with gallery size equals to 100.}
	\label{tab2}	
\end{table}

\subsubsection{Impact of Gallery Size}

\begin{figure}[ht!]
	\centering
	\subfigure[]{
		\includegraphics[width=0.7\linewidth]{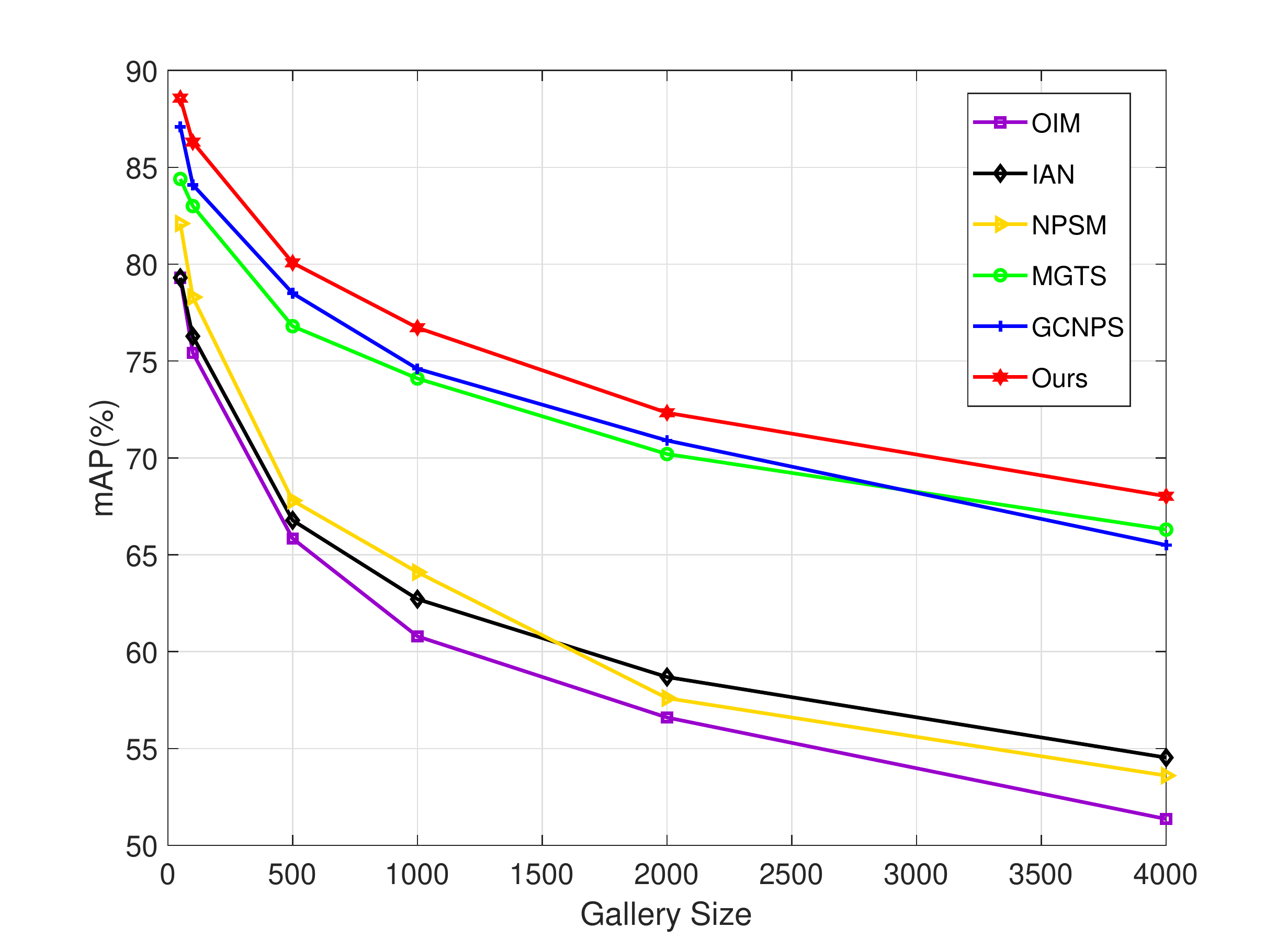}}\\
	\subfigure[]{
		\includegraphics[width=0.7\linewidth]{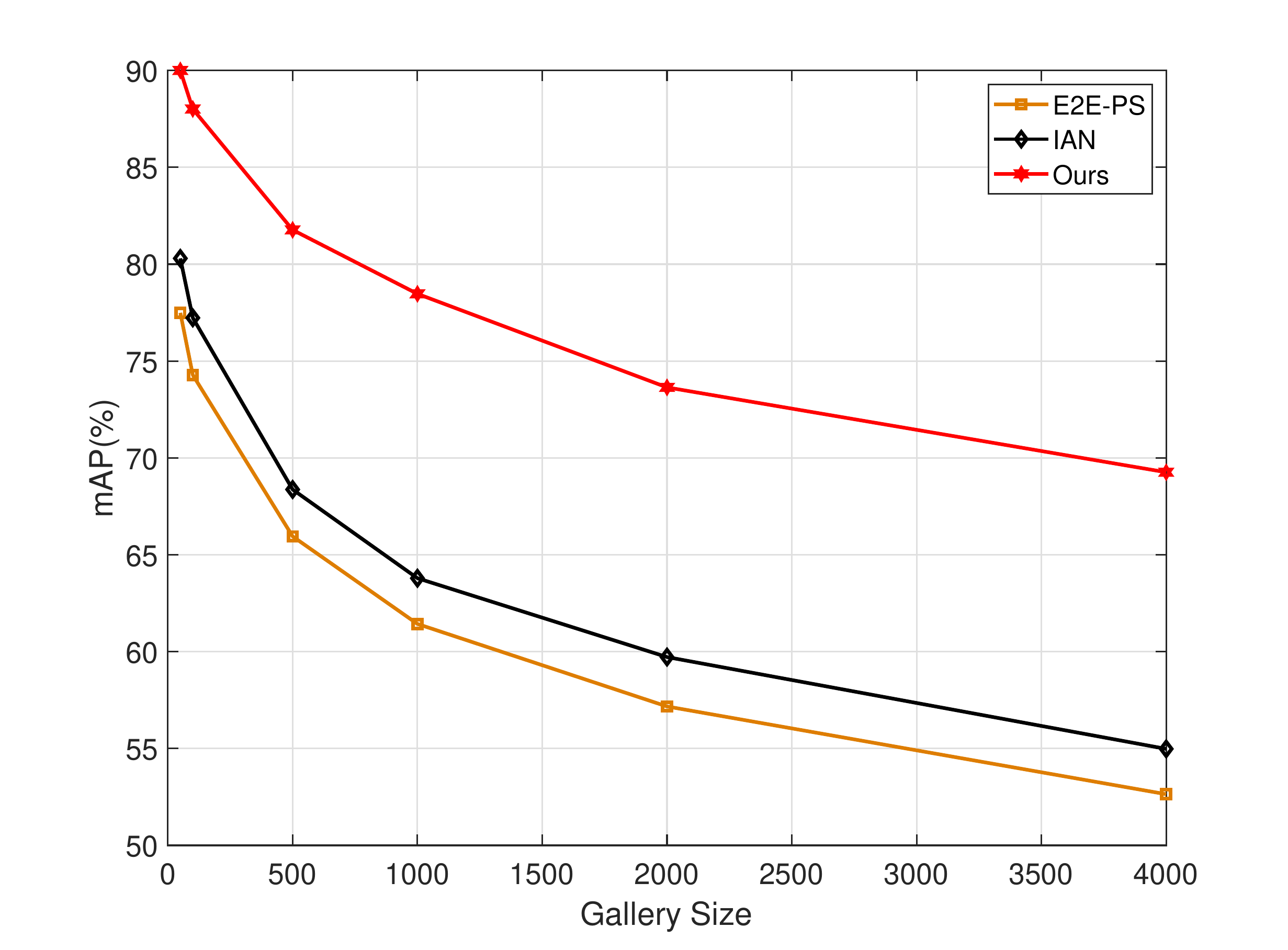}}
	
	\setlength{\abovecaptionskip}{10pt}%
	\caption{Person search performance comparison on CUHK-SYSU dataset with different gallery size, [50, 100, 500, 1,000, 2,000, 4,000]. (a) Model with ResNet-50 backbone. (b) Model with ResNet-101 backbone.}
	
	\label{fig4}
\end{figure}

Each gallery image in CUHK-SYSU dataset contains around $6$ pedestrians on average. With gallery size $100$, person search aims to retrieve each target person from about $600$ pedestrians. The person search is more challenging with the increasing number of gallery size. We also report the performance of our model with various gallery size, including [$50$, $100$, $500$, $1,000$, $2,000$, $4,000$]. The results are demonstrated in Fig. \ref{fig4}. As expected, the person search performance of all methods drops with the increasing gallery size. While our person search framework remains superior than other approaches with various gallery sizes.  

\subsubsection{Impact of Occlusion and Low Resolution} 

Person search becomes even harder when pedestrians are occluded or the resolution of the captured images is low. Therefore, to prove the robustness of our method, we further evaluate our model on two subsets. One subset contains $187$ target persons with occlusion, and the other subset contains $290$ target persons with low resolution. The results are demonstrated in Table \ref{tab3}. We follow the notations defined in \cite{xiao2019ian}, where ``whole" denote the full set which contains 2,900 probe images. We observe that the performance degenerate under these two extreme conditions compared with full set. However, our person search framework still outperforms the other approaches \cite{xiao2016end}\cite{xiao2019ian}.



\begin{table}
	\centering
	\begin{tabular}{lcccccc}
		\hline 
		\multirow{2}{*}{\textbf{Method}}  &  \multicolumn{2}{c}{Low-Res} & \multicolumn{2}{c}{Occulusion} & \multicolumn{2}{c}{Whole} \\
		\cmidrule(lr){2-3}\cmidrule(lr){4-5}\cmidrule(lr){6-7}
		& mAP(\%) & top-1(\%)  
		& mAP(\%) & top-1(\%)
		& mAP(\%) & top-1(\%) \\
		\hline 
		E2E-PS(VGGNet)  & 46.1 & 51.0 & 44.3 & 45.4 & 69.6 & 72.9 \\
		E2E-PS(Res-101)  & 47.9 & 52.0 & 47.7 & 48.1 & 74.2 & 78.1 \\
		IAN(Res-101) & 52.6 & 54.4 & 53.0 & 54.5 & 77.2 & 80.4 \\
		Ours(Res-50)     & \textbf{66.7} & \textbf{66.8} & \textbf{70.8} & \textbf{71.3} & \textbf{86.3}  & \textbf{86.5} \\
		\hline
		
	\end{tabular}
	\setlength{\abovecaptionskip}{10pt}%
	\caption{Person search performance on low resolution and occlusion subsets.}
	\label{tab3}
\end{table}

\subsection{Ablation Study}

With the assist of the newly labeled dataset, our proposed person search framework produces more discriminative features by utilizing partially labeled segmentation mask. To evaluate the effectiveness of our approach, we report the person search performance when we progressively increase the number of images with segmentation mask. The results are shown in Table \ref{tab4}, where we denote the proportion of the images with segmentation mask as $\alpha$. In total, 1,833 images are labeled with segmentation mask, which accounts for around 16\% of the 11,206 training images. When all those 1,833 images are used for training, we denote as ``Full". It can be observed that there is an obvious gain when 12\% images with segmentation mask are used for training, and tend to be stable until 15\% images are used. That is why we only label 16\% of all the images. 

\begin{table}[h]
	\begin{center}
		\begin{tabular}{p{2cm}<{\centering}|p{1cm}<{\centering} p{1cm}<{\centering} p{1cm}<{\centering} p{1cm}<{\centering} p{1cm}<{\centering} p{1cm}<{\centering}}
			\hline
			Value of $\alpha$ & 3\% & 6\% & 9\% & 12\% & 15\% & Full \\ 
			\hline
			mAP(\%) & 85.1 & 85.3 & 85.3 & 86.1 & \textbf{86.3} & \textbf{86.3} \\
			top-1(\%) & 85.2 & 85.4 & 85.7 & 86.1 & \textbf{86.5} & \textbf{86.5} \\
			
			\hline
		\end{tabular}
		
		\setlength{\abovecaptionskip}{10pt}%
		\caption{Person search performance on CUHK-SYSU dataset with various proportion of images with segmentation masks}
		\label{tab4}
	\end{center}
\end{table}

\begin{figure}[ht!]
	\centering
	\subfigure[]{
		\includegraphics[width=1 \linewidth]{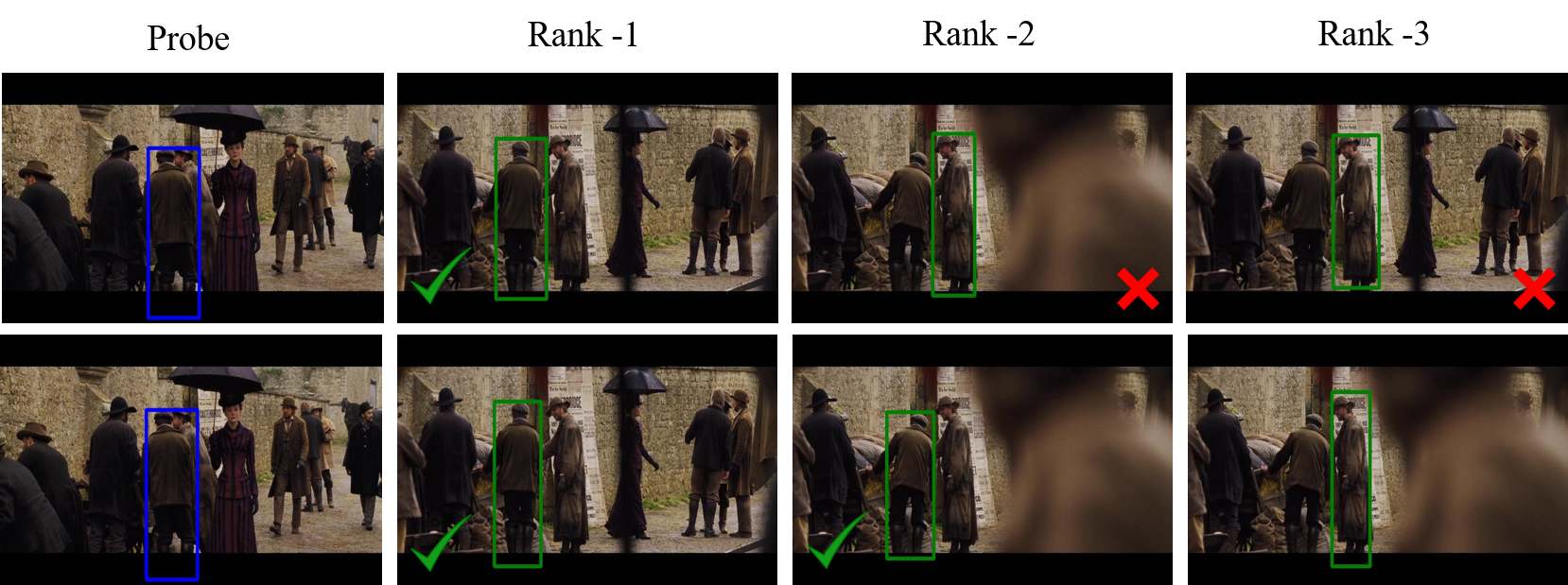}}\\
	
	\subfigure[]{
		\includegraphics[width=1 \linewidth]{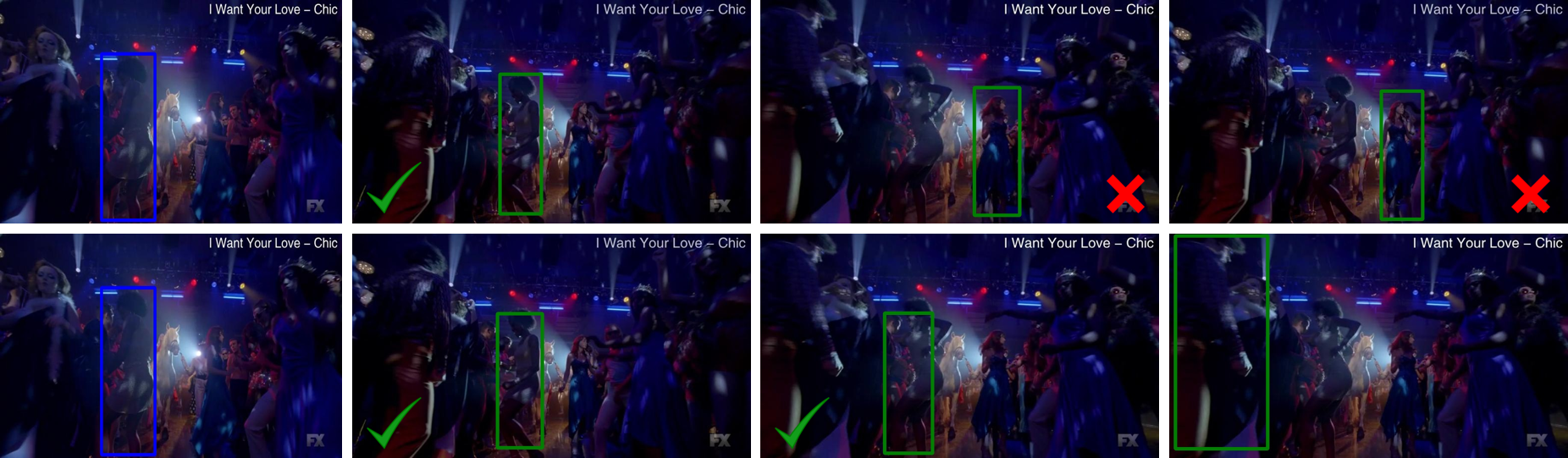}}\\
	
	\subfigure[]{
		\includegraphics[width=1 \linewidth]{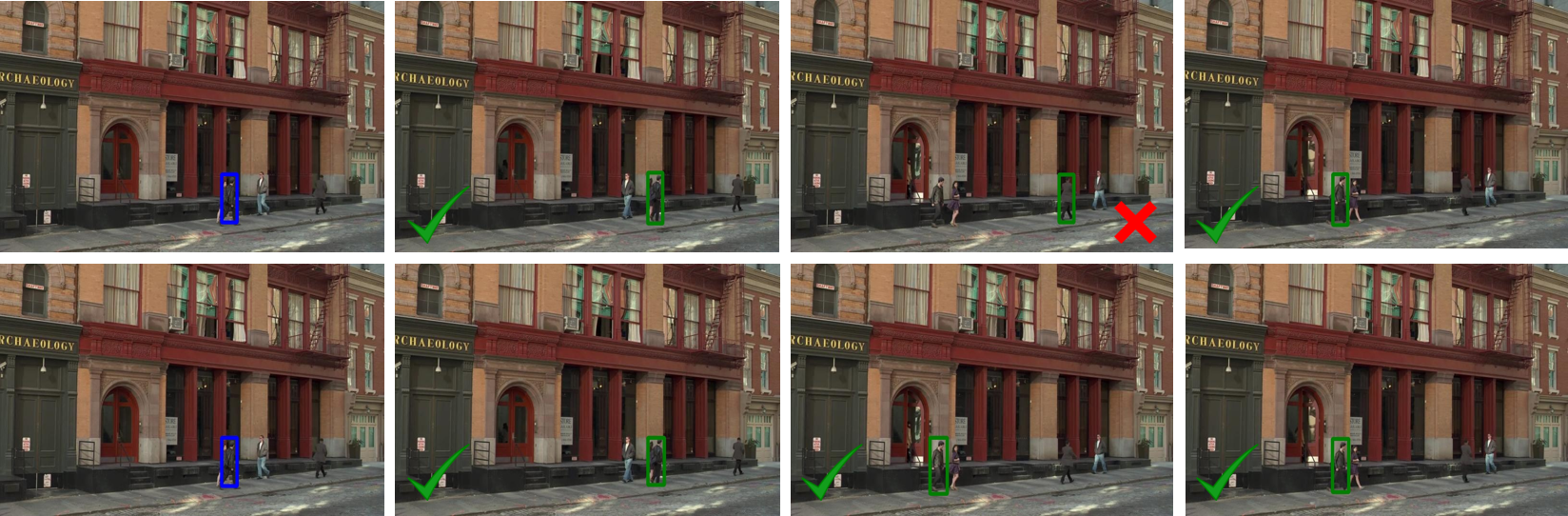}}\\

	\setlength{\abovecaptionskip}{10pt}%
	\caption{Three groups of top-3 comparison results for person search on CHUK-SYSU dataset. The upper row in each group are the searching results of OIM \cite{xiao2017joint}, and the lower row in each group are the searching results of our model (both models adopt ResNet-50 as backbone). The blue boxes in the first column indicate the probe images, and the green boxes in other columns indicate their top-3 searching results. Best viewed in color.}
	
	\label{fig5}
\end{figure}

\section{Conclusion}

Person search handles the challenges from both pedestrian detection and person identification, and inevitably introduces background clutters into the detected candidate boxes. To address this problem, with the assist of our new created dataset which contains the labeled segmentation masks for a portion of images in the existing CUHK-SYSU dataset, we propose a novel segmentation mask guided person search framework to extract more discriminative and robust features invariant to background clutters for each human individual. Moreover, our person search framework is trained end-to-end, which proves that joint optimization of pedestrian detection, person re-identification, and pedestrian segmentation is an effective solution for person search. Finally, extensive experiments show that our proposed method achieves state-of-the-art performance on CUHK-SYSU dataset.

\bibliography{mybibtex}
\end{document}